\documentclass[runningheads]{llncs}
\usepackage{graphicx}
\usepackage{amsmath,amssymb} 
\usepackage{color}
\usepackage{bm}
\usepackage{float} 
\usepackage{caption}
\usepackage{subcaption}
\usepackage{multirow}
\usepackage{booktabs}

\usepackage[width=122mm,left=12mm,paperwidth=146mm,height=193mm,top=12mm,paperheight=217mm]{geometry}
\usepackage{cite}
\usepackage{hyperref}
\usepackage{xcolor}
\usepackage{url}

\usepackage{algorithm}
\usepackage{algorithmicx}
\usepackage{algpseudocode}
\usepackage{amsmath} 
\floatname{algorithm}{Algorithm} 

\let\llncssubparagraph\subparagraph
\let\subparagraph\paragraph
\usepackage[compact]{titlesec}
\let\subparagraph\llncssubparagraph
\titlespacing*{\section}{0pt}{2.5ex plus 1ex minus .0ex}{2.0ex plus .0ex}
\titlespacing*{\subsection}{0pt}{1.5ex plus 1ex minus .0ex}{1.5ex plus .0ex}
\titlespacing*{\subsubsection}{0pt}{0.1ex plus 1ex minus .0ex}{0.1ex plus .0ex}
\hypersetup{
  colorlinks=true,
  linkcolor=green!75!black,
  citecolor=green!75!black,
  urlcolor=blue,  
}

\begin{document}
\pagestyle{headings}

\mainmatter
\def\ECCV24SubNumber{***}  

\title{PSCR: Patches Sampling-based Contrastive Regression for AIGC Image Quality Assessment} 

\titlerunning{PSCR: Patches Sampling-based Contrastive Regression for AIGC Image Quality Assessment}
\authorrunning{J. Yuan, X. Cao, L. Cao, J. Lin, X. Cao}

\author{Jiquan Yuan, Xinyan Cao, Linjing Cao, \\Jinlong Lin, \and
Xixin Cao\thanks{Corresponding author. Email: cxx@ss.pku.edu.cn}}
\institute{School of Software \& Microelectronics, Peking University, Beijing, China}

\maketitle

\begin{abstract}
In recent years, Artificial Intelligence Generated Content (AIGC) has gained widespread attention beyond the computer science community. Due to various issues arising from continuous creation of AI-generated images (AIGI), AIGC image quality assessment (AIGCIQA), which aims to evaluate the quality of AIGIs from human perception perspectives, has emerged as a novel topic in the field of computer vision. However, most existing AIGCIQA methods directly regress predicted scores from a single generated image, overlooking the inherent differences among AIGIs and scores. Additionally, operations like resizing and cropping may cause global geometric distortions and information loss, thus limiting the performance of models. To address these issues, we propose a patches sampling-based contrastive regression (PSCR) framework. We suggest introducing a contrastive regression framework to leverage differences among various generated images for learning a better representation space. In this space, differences and score rankings among images can be measured by their relative scores. By selecting exemplar AIGIs as references, we also overcome the limitations of previous models that could not utilize reference images on the no-reference image databases. To avoid geometric distortions and information loss in image inputs, we further propose a patches sampling strategy. To demonstrate the effectiveness of our proposed PSCR framework, we conduct extensive experiments on three mainstream AIGCIQA databases including AGIQA-1K, AGIQA-3K and AIGCIQA2023. The results show significant improvements in model performance with the introduction of our proposed PSCR framework. Code will be available at \url{https://github.com/jiquan123/PSCR}.
\end{abstract}

\section{Introduction}
In recent years, Artificial Intelligence Generated Content (AIGC) has garnered extensive attention beyond the field of computer science. AIGC, created by advanced Generative AI (GAI) technologies rather than human authors, can automatically produce large volumes of content in a short time. Its core concept involves using AI generative models to automatically create various types of content such as text, images, audio, and video, based on given themes, keywords, formats, and styles, \textit{etc.} AIGC is widely applicable in fields like media, education, entertainment, marketing, and scientific research,  \textit{etc.}, offering users high-quality, efficient, and highly personalized content services.

\begin{figure*}[t]
\centering
	\subcaptionbox{}{\includegraphics[width = 1.41cm]{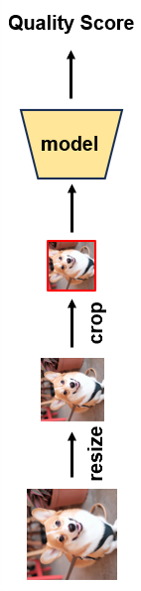}}
	\hfill
	\subcaptionbox{}{\includegraphics[width = 2.60cm]{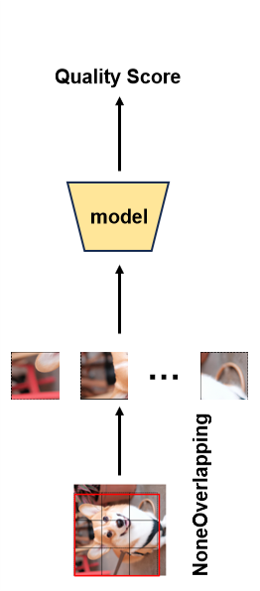}}
	\hfill
	\subcaptionbox{}{\includegraphics[width = 4.55cm]{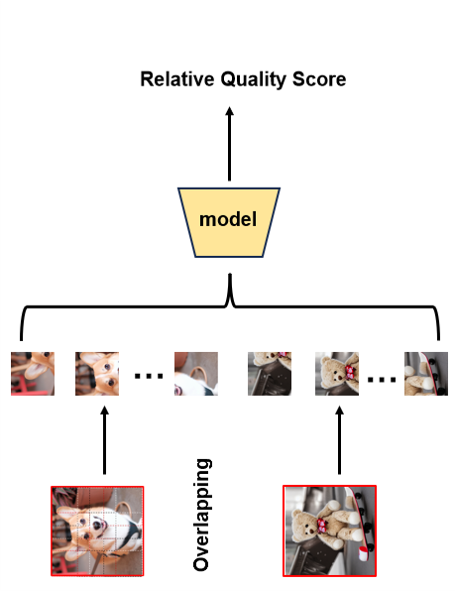}} 
\caption{Most IQA methods employ a direct regression approach to obtain predicted scores such as (a) and (b). The image preprocessing methods they use, such as resizing, cropping, and non-overlapping patches sampling, often lead to geometric distortion and information loss. Our proposed patches sampling-based contrastive regression framework (c) initially captures overlapping image patches using a sliding window to avoid information loss, and simultaneously learns a better representation space by leveraging differences among various images. The red box section within the image represents the image information utilized.}
\label{fig:label}
\end{figure*}

 Due to various issues arising from continuous creation of AI-generated images (AIGI), AIGC image quality assessment (AIGCIQA), which aims to evaluate the quality of AI-generated images from human perception perspectives, has emerged as a novel topic in the field of computer vision. Compared to common image content, AIGIs often suffer from unique distortions\cite{zhang2023perceptual,wang2023aigciqa2023}, such as unrealistic structures, irregular textures and shapes, and AI artifacts, \textit{etc.}, making AIGCIQA more challenging. Unlike traditional image classification or object detection tasks, which mainly focus on classifying and segmenting objects in images, AIGCIQA requires models to predict fine-grained scores from a multitude of AIGIs. Considering the variations in AIGIs and scores, the key to addressing this issue lies in enabling models to learn the differences among AIGIs and regress predicted scores based on these differences.

 Over the past few years, significant efforts have been made to advance the development of AIGCIQA, such as the establishment of dedicated databases like AGIQA-1K\cite{zhang2023perceptual}, AGIQA-3K\cite{AGIQA-3K}, AIGCIQA2023\cite{wang2023aigciqa2023}, and PKU-I2IQA\cite{yuan2023pkui2iqa}, \textit{etc.} However, we identify two main deficiencies in current research on AIGCIQA: First, most of them\cite{cnriqa,iqa1,iqa2, iqa3,zhang2023perceptual, AGIQA-3K, wang2023aigciqa2023, yuan2023pkui2iqa} treat AIGCIQA as a regression problem, where models directly regress predicted scores from individual AIGIs, overlooking the inherent differences among AIGIs and their scores. Second, the majority of AIGCIQA methods\cite{cnriqa,iqa1,iqa2, iqa3,zhang2023perceptual, wang2023aigciqa2023, yuan2023pkui2iqa} resize, crop, or otherwise preprocess before inputting them into the model, potentially causing global geometric distortion and critical information loss, thereby impacting model performance. In some traditional computer vision tasks, input images may contain both foreground and background information. In these tasks, preprocessing like resizing or cropping typically discards background information, generally not affecting model performance. However, in AIGCIQA tasks, which require predicting fine-grained scores from AIGIs, every piece of information in the AIGIs is crucial. To address these issues in existing AIGCIQA methods, we propose a patches sampling-based contrastive regression (PSCR) framework.

Addressing the first issue mentioned earlier, to better utilize the inherent differences between AIGIs for score prediction, we introduce a contrastive regression framework\cite{yu2021core, zha2022rankCR}. Unlike previous AIGCIQA methods, whose learned representations are often dispersed and fail to capture the continuous nature of regression tasks, contrastive regression aims to map input images to a representation space where differences among images and their score rankings can be measured by their relative scores. Specifically, we randomly select some exemplar AIGIs as references for the input query AIGIs, and then regress the relative scores between the query AIGIs and these exemplar AIGIs. This framework not only leverages differences among various inputs to learn a better representation space but also overcomes the limitation of previous no-reference image quality assessment (NR-IQA) methods by using exemplar AIGIs as references.
Regarding the second issue, we propose a patches sampling strategy. To avoid geometric distortion and information loss in AIGIs before they are fed into the model, we employ a sliding window on the original resolution AIGIs to obtain overlapping image patches of a specific size. The differences between our proposed method and the majority of existing methods are illustrated in Fig.1.

To demonstrate the effectiveness of our proposed method, we conduct extensive experiments on three mainstream AIGCIQA databases containing AIGIs generated by various models such as Midjourney\cite{midjourney}, Stable Diffusion\cite{r2}, \textit{etc.}, including AGIQA-1K\cite{zhang2023perceptual}, AGIQA-3K\cite{AGIQA-3K}, and AIGCIQA2023\cite{wang2023aigciqa2023}. We adopt the no-reference image quality assessment method NR-AIGCIQA proposed in \cite{yuan2023pkui2iqa} as our baseline model. The results indicate significant performance improvement by incorporating our proposed PSCR framework. Our contributions are summarized as follows:

$\bullet$  We propose a patches sampling-based contrastive regression framework. As a versatile framework, it can be conveniently integrated into existing IQA methods.

$\bullet$  We conduct extensive experiments on three mainstream AIGCIQA databases to demonstrate the effectiveness of our proposed PSCR framework.

\section{Related Work}
\subsubsection{Image Quality Assessment.}
Over the past few years, researchers have proposed numerous Image Quality Assessment (IQA) methods. Mainstream IQA methods often treat IQA as a regression problem. Initially, many classic image quality assessment models utilize hand-crafted feature-based methods\cite{nr-iqa,r19,r20}. However, with the rapid development of convolutional neural networks, deep learning-based feature extraction methods\cite{cnriqa,iqa1,kim,Li,pan,yan,iqa2, iqa3} have significantly improved performance. In the field of no-reference image quality assessment (NR-IQA), Kang \textit{et al.}\cite{cnriqa} propose using image patches as inputs for CNNs, obtaining quality scores for each patch at the output nodes, and averaging these scores to determine the overall image quality score. They employ the image preprocessing operation depicted in Fig.1(b), segmenting the image into non-overlapping $32\times32$ patches. Kim \textit{et al.}\cite{kim} introduce a deep no-reference image quality assessment model based on CNNs, utilizing local quality maps obtained from full-reference image quality assessment algorithms as intermediate regression targets to address the lack of ground-truth values for image patches. This approach also involve training by merging image patches to assess overall image quality. Yan \textit{et al.}\cite{yan} propose a dual-stream convolutional network [13], which enhances performance by utilizing information from multiple channels, reducing the difficulty of extracting features from single-stream CNNs. Pan \textit{et al.}\cite{pan} introduce an algorithm based on multi-branch CNNs and propose a weighting mechanism considering the positional relationship between image patches and the entire image to improve the accuracy of final quality prediction. Li \textit{et al.}\cite{Li} suggest a no-reference image quality assessment algorithm based on multi-scale feature representation. To avoid overfitting, they employ CNN-based methods to crop large images into patches and use random horizontal flipping and transfer learning for data augmentation. 
As a branch of IQA, research in the field of AIGCIQA remains in its nascent stages. Several dedicated AIGCIQA databases have been proposed to promote AIGCIQA development, such as AGIQA-1K\cite{zhang2023perceptual}, AGIQA-3K\cite{AGIQA-3K}, AIGCIQA2023\cite{wang2023aigciqa2023}, and PKU-I2IQA\cite{yuan2023pkui2iqa}, \textit{etc.} These studies mostly use some of the current IQA models\cite{iqa1, he2016resnet, iqa2, cnriqa, iqa3, simonyan2014vgg} for benchmark experiments. Li \textit{et al.}\cite{AGIQA-3K} propose StairReward, significantly enhancing subjective text-to-image alignment evaluation performance. Yuan \textit{et al.}\cite{yuan2023pkui2iqa} introduce two benchmark models NR-AIGCIQA for no-reference image quality assessment and FR-AIGCIQA for full-reference image quality assessment. However, these methods mostly regress scores directly from single generated images and often do not utilize other images in the database for comparative reference. Additionally, the preprocessing operations on images by these methods may lead to geometric distortion and information loss, thus limiting the models' performance. In this work, we propose a patches sampling-based contrastive regression framework, aiming to avoid geometric distortions and information loss in image inputs, while utilizing differences between inputs to learn a better representation space.
\subsubsection{Contrastive Regression.}
As one of the most common and fundamental problems in the real world, regression problems are involved in various fields and computer vision tasks, such as predicting athletes' action quality scores\cite{parmar2017learning}, predicting age from human appearance\cite{agedb}, and forecasting temperature from outdoor cameras\cite{weather}, \textit{etc.} Existing regression models mostly regress prediction values directly in an end-to-end manner. Yu \textit{et al.}\cite{yu2021core} introduce a contrastive regression (CoRe) framework in the AQA task, regressing relative scores through pairwise comparisons, highlighting differences among videos. More recently, Zha \textit{et al.}\cite{zha2022rankCR} propose the Rank-N-Contrast framework, which contrasts samples based on their ranking in the target space to learn continuous representations for regression. In our work, we combine the contrastive regression framework with our proposed patches sampling strategy and propose a patches sampling-based contrastive regression framework for AIGCIQA. Extensive experiments on several mainstream AIGCIQA databases demonstrate the effectiveness of our proposed method.

\begin{figure}[t]
\centering
\includegraphics[width=11.26cm,height=5.13cm]{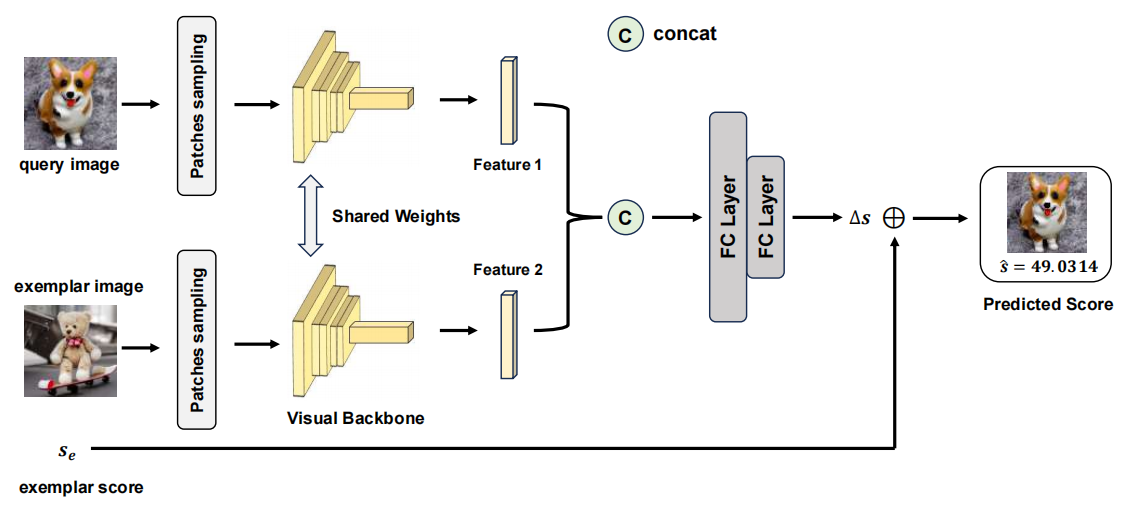} \label{1}
\caption{The pipeline of our proposed patches sampling-based contrastive regression framework. We first randomly sample an exemplar image for each query image. We then use a sliding window to obtain image patches and feed the pairwise image patches into a shared-weights visual backbone to extract features and fuse these two features by concatenation. Finally, we pass the fused feature to the score regression network composed of two fully connected layers and obtain the score difference between the two images. During inference, the final score can be computed by averaging the results from multiple different exemplars.}
\end{figure}

\section{Approach}
The overall framework of our method is illustrated in Fig.2. We will describe our method in detail as follows.

\subsection{Problem Formulation.}
Given pairwise query images $I_q$ and exemplar images $I_e$ with score labels $s_q$ and $s_e$, we first divide each image into $N$ overlapping image patches using a sliding window. These pairwise patches are then fed into a shared-weights visual backbone network for feature extraction, followed by a concatenation process to fuse these features. Then, we employ average pooling to aggregate image patches features for capturing spatial context relationships. Finally, a regression network composed of two fully connected layers utilizes the fused features to regress the relative scores between the query and exemplar images. It can be represented as:
\begin{align}
   \Delta \hat{s} = R_\theta (F_w ({PS}(I_q)), F_w ({PS}(I_e)))  \label{Eq.4} 
\end{align}
where $\Delta \hat{s}$ is the predicted relative score between the query image $I_q$ and the example image $I_e$, $R_\theta$ and $F_w$ represent the regression network and feature extraction network with parameters $_\theta$ and $w$ respectively, and ${PS}$ denotes patches sampling.

\subsection{Patches Sampling.}
Given an AIGI with a resolution of $512\times512$, our proposed patches sampling strategy utilizes a sliding window to divide the image into $N$ specified-size overlapping image patches. Specifically, we utilize sliding windows of different strides and sizes for various visual backbone networks. The start-index of the sliding window is used to indicate its initial position. For InceptionV4, we select a list of start-indices as $[0, 100, 213]$ with a window size of $299\times299$; for models like VGG and ResNet, our chosen list of start-indices is $[0, 150, 288]$ with a window size of $224\times224$. The patches sampling process is summarized in Algorithm 1.

\begin{algorithm}[t]
    \caption{Overlapping Patches Sampling with Sliding Window}
    \begin{algorithmic}[1]
        \Require The list of start-indices $S$, the size of sliding window $s$, image $I$
        \Ensure The list of sampled image patches $I_p$
        \State Initialize $I_p$ = []
        \For{$i$ in $S$}
            \For{$j$ in $S$}
                \State Append $I[:, :, i:(i + s), j:(j + s)]$ to $I_p$
            \EndFor
        \EndFor             
    \end{algorithmic}
\end{algorithm}

\subsection{Contrastive Regression.}
Most existing AIGCIQA methods\cite{cnriqa, iqa1, iqa2, iqa3, yuan2023pkui2iqa} frame the task as a regression problem, directly regressing predicted scores from individual AIGIs, overlooking the inherent differences among AIGIs and their scores. These methods often learn dispersed representations, failing to capture the continuous nature of regression tasks, as well as the order and differences among regression scores. Formally, for a given AIGI $I_g$ with a score label $s$, a feature extraction network is first used to extract features from $I_g$, followed by a regression network to obtain the predicted score. This can be expressed as:
\begin{align}
   \hat{s} = R_\theta(F_w(I_g)) \label{Eq.4} 
\end{align}

where $R_\theta$ and $F_w$ respectively denote the regression network with parameters $_\theta$ and the feature extraction network with parameters $w$. These methods typically optimize the parameters of the feature extraction network and the score regression network by minimizing the mean squared error between the predicted score $\hat{s}$ and the ground-truth score $s$:
\begin{align}
   L(\theta, w | I) = {MSE}(\hat{s}, s) \label{Eq.4} 
\end{align}
Here, $_\theta$ and $w$ are the parameters of the regression network and feature extraction network, respectively.

However, such direct regression methods often struggle to enable models to learn the inherent differences among AIGIs and their scores. In contrast, contrastive regression\cite{yu2021core, zha2022rankCR} aims to map input images to a representation space where differences among images and their score rankings can be measured by their relative scores. Therefore, we suggest introducing a contrastive regression framework to better utilize differences among AIGIs for regression-based score prediction. Specifically, for a given query AIGI, we randomly select some exemplar AIGIs as references and then regress the relative scores between the query AIGIs and the exemplar AIGIs. As previously described, given pairwise query images $I_q$ and exemplar images $I_e$ with score labels $s_q$ and $s_e$, the regression problem can be reformulated as:

\begin{align}
   \Delta \hat{s} = R_\theta(F_w(I_q), F_w(I_e)) \label{Eq.4} 
\end{align}

where $\Delta \hat{s}$ is the predicted relative score between the query image $I_q$ and the example image $I_e$. The predicted score for the query AIGI is then calculated by the following formula:

\begin{align}
   \hat{s}_q = \Delta \hat{s} + s_e \label{Eq.4} 
\end{align}

Accordingly, we optimize the parameters of the feature extraction networks and regression networks by minimizing the mean squared error between the predicted and ground-truth scores:

\begin{align}
   L(\theta, w | I) = {MSE}(\hat{s}_q, s_q) = {MSE}(\Delta \hat{s}, s_q - s_e) \label{Eq.4} 
\end{align}

The contrastive regression framework not only leverages the differences among various inputs to learn a better representation space but also addresses the limitation of previous NR-IQA methods that could not utilize reference images by selecting example AIGIs as references.

\subsection{Inference.}
During testing, for a test image $I_{test}$, we adopt a multi-exemplar voting strategy\cite{yu2021core} to select $N$ exemplars from the training set and then construct $N$ image pairs $\{I_{test},I_j\}_{j=1}^N$ with exemplar score labels $\{s_{I_j}\}_{j=1}^N$. The process of multi-exemplar voting can be written as:

\begin{align}
   \hat{s}_{I_{test}} = \frac{1}{N} \sum_{j=1}^{N} \left( R_{\theta}(F_w({PS}(I_{test})), F_w({PS}(I_j))) + s_{I_j} \right)
 \label{Eq.4} 
\end{align}

Here, $\hat{s}_{I_{test}}$ represents the predicted score for the test image $I_{test}$.

\begin{figure*}[t]
\centering
	\subcaptionbox{}{\includegraphics[width = 6.0cm]{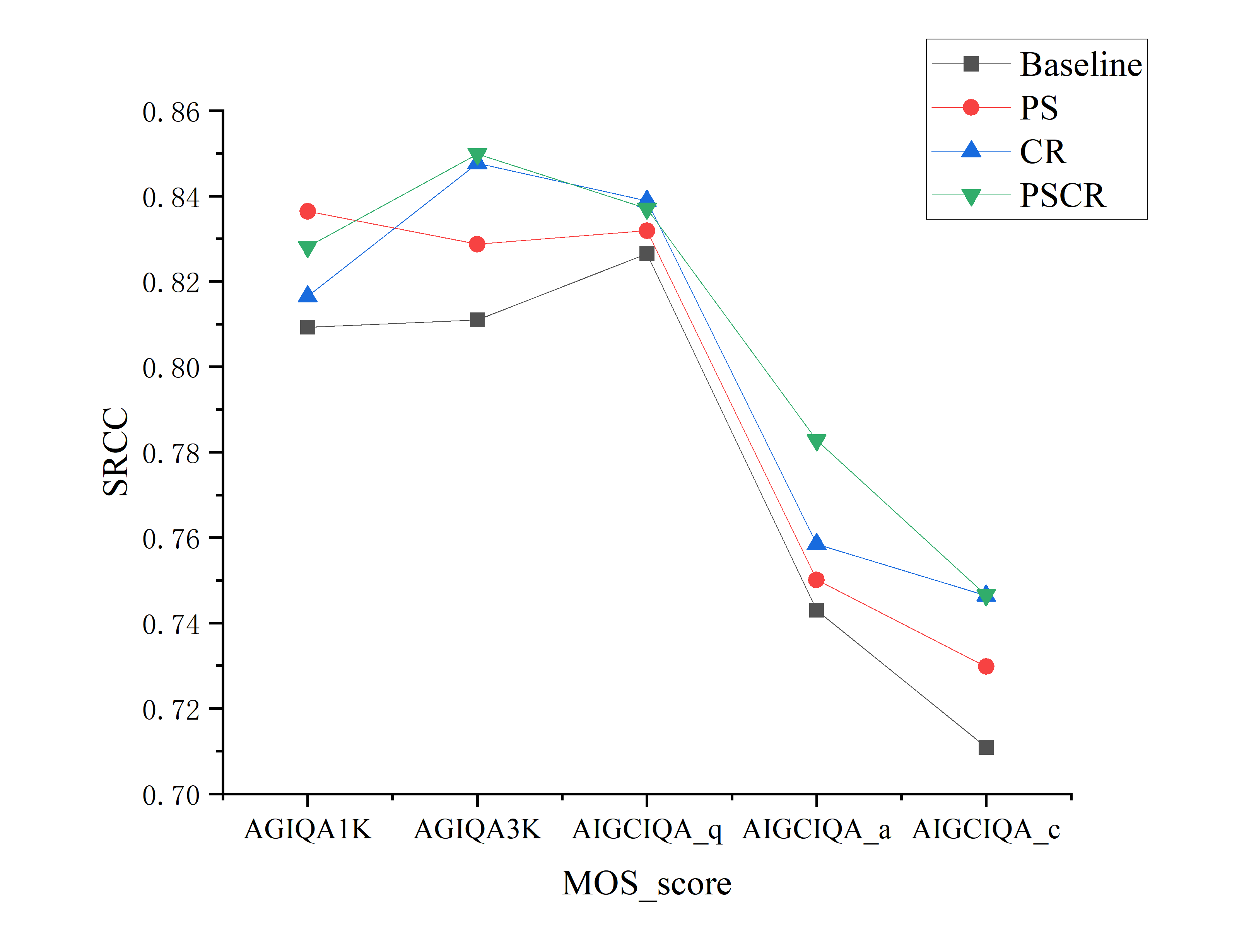}}
	\hfill
	\subcaptionbox{}{\includegraphics[width = 6.0cm]{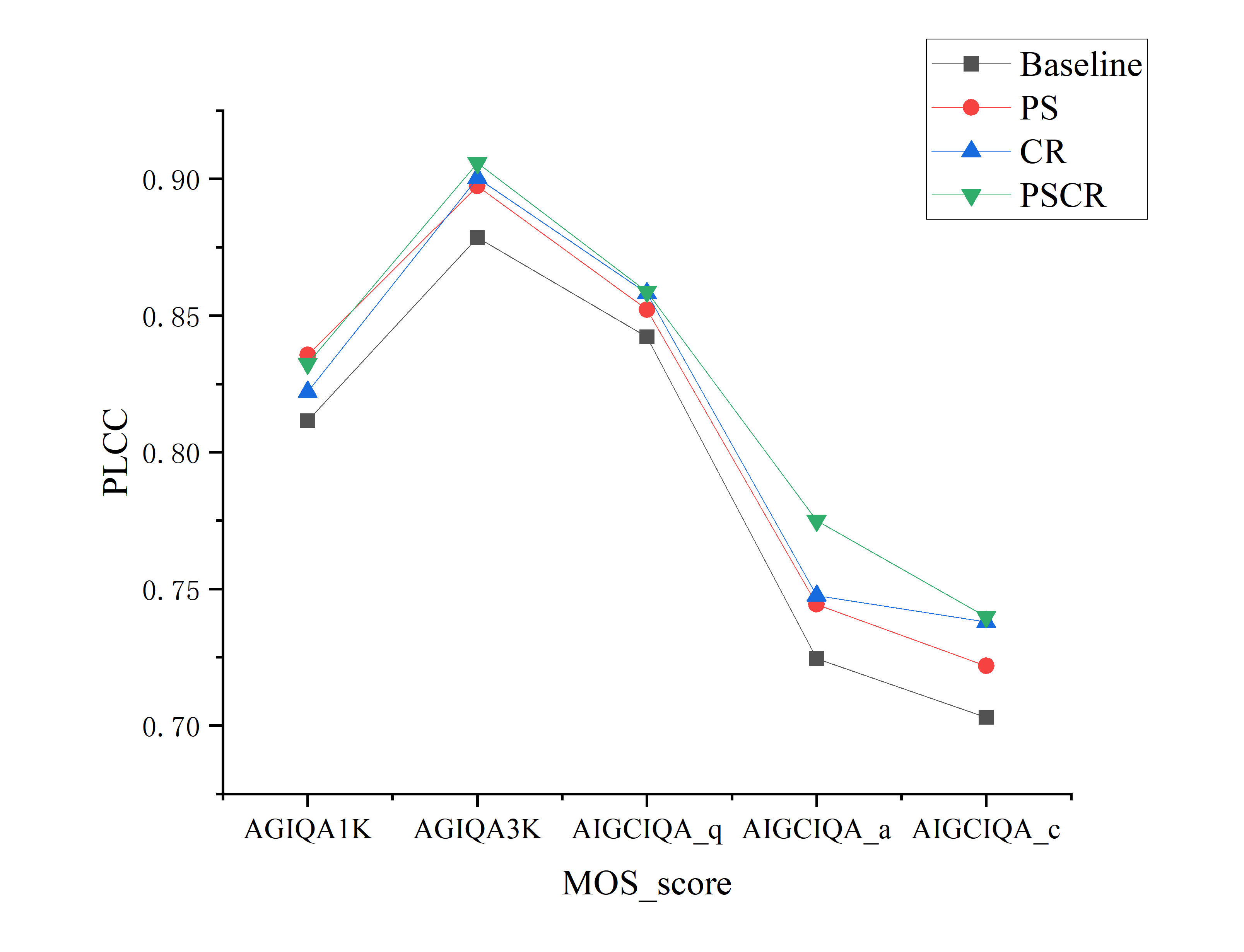}}
	 
\caption{Comparison of the performance between our proposed methods and the Baseline on three mainstream AIGCIQA database under SRCC and PLCC. As an example, it demonstrates the performance while employing InceptionV4 for feature extraction. }
\label{fig:label}
\end{figure*}

\section{Experiment}

\subsection{Datasets.}
We perform experiments on three mainstream AIGCIQA benchmarks including AGIQA-1K\cite{zhang2023perceptual}, AGIQA-3K\cite{AGIQA-3K} and AIGCIQA2023\cite{wang2023aigciqa2023}.

\subsubsection{AGIQA-1K.}
The AGIQA-1K database contains 1080 AIGIs generated by two Text-to-Image models\cite{r2} stable-inpainting-v1 and stable-diffusion-v2. To ensure content diversity and catch up with the popular trends, the authors use the hot keywords from the PNGIMG website for AIGIs generation. The generated images are of bird, cat, Batman, kid, man, and woman, \textit{etc.}

\subsubsection{AGIQA-3K.}
The AGIQA-3K database contains 2982 AIGIs generated by six Text-to-Image models including GLIDE\cite{nichol2021glide}, Stable Diffusion V1.5\cite{r2}, Stable Diffusion XL2.2\cite{SDXL22}, Midjourney\cite{midjourney}, AttnGAN\cite{xu2018attngan} and DALLE2\cite{dalle}. This is the first database that covers AIGIs from GAN/auto regression/diffusion-based model altogether.

\subsubsection{AIGCIQA2023.}
The AIGCIQA2023 database contains 2400 AI-generated images generated by six of the latest Text-to-Image models including Glide\cite{nichol2021glide}, Lafite\cite{zhou2022lafite}, DALLE\cite{dalle}, Stable-diffusion\cite{r2}, Unidiffusion\cite{bao2023unidiff}, Controlnet\cite{zhang2023addingcontrolnet}. They gather 100 text prompts from PartiPrompts (10 different scene categories $\times$ 10 challenge categories) and generate four different images for each model randomly. Therefore, the constructed AIGCIQA-2023 database contains a total of 2400 AIGI (4 images $\times$ 6 models $\times$ 100 prompts) corresponding to 100 prompts.

\subsection{Evaluation Criteria}
Following prior research \cite{zhang2023perceptual,wang2023aigciqa2023, AGIQA-3K, yuan2023pkui2iqa}, we utilize the Spearman rank correlation coefficient (SRCC) and Pearson linear correlation coefficient (PLCC) as evaluation metrics to evaluate the performance of our model.

The SRCC is defined as follows:
\begin{align}
   \text{SRCC} = 1 - \frac{6 \sum_{i=1}^{N} d_i^2}{N(N^2 - 1)} \label{Eq.4} 
\end{align}

Here, $N$ represents the number of test images, and $d_i$ denotes the difference in ranking between the true quality scores and the predicted quality scores for the $i_{th}$ test image.

The PLCC is defined as follows:
\begin{align}
    \text{PLCC} = \frac{\sum_{i=1}^{N}(si - \mu_{s_i})(\hat{s}_i - \hat{\mu}_{s_i})}{\sqrt{\sum_{i=1}^{N}(s_i - \mu_{s_i})^2 \sum_{i=1}^{N}(\hat{s}_i - \hat{\mu}_{s_i})^2}}\label{Eq.4} 
\end{align}

Here, $s_i$ and $\hat{s}_i$ represent the true and predicted quality scores, respectively, for the $i_{th}$ image. $\mu_{s_i}$ and $\hat{\mu}_{s_i}$ are their respective means, and $N$ is the number of test images.
Both SRCC and PLCC are metrics used to evaluate the relationship between two sets of variables. They range between $-1$ and $1$, where a positive value indicates a positive correlation and a negative value indicates a negative correlation, and a larger value means a better performance.

\subsection{Implementation Details}
Our experiments are conducted on the NVIDIA A40, using PyTorch 1.11.0 and CUDA 11.3 for both training and testing.

For feature extraction from input images, we select several backbone network models pre-trained on the ImageNet\cite{russakovsky2015imagenet}, including VGG16\cite{simonyan2014vgg}, VGG19\cite{simonyan2014vgg}, ResNet18\cite{he2016resnet}, ResNet50\cite{he2016resnet}, and InceptionV4\cite{szegedy2017inception}. For InceptionV4, the start-indices list we choose during patches sampling is [0, 100, 213] with a sliding window size of $299 \times 299$. For models such as VGG and ResNet, the selected start-indices list is [0, 150, 288] with a sliding window size of $224 \times 224$. During training, the batch size $B$ is set to $8$. We utilize the Adam optimizer\cite{kingma2014adam} with a learning rate of $1 \times 10^{-4}$ and weight decay of $1 \times 10^{-5}$. The training loss employed is mean squared error (MSE) loss. In the testing phase, the batch size $B$ is set to $20$. We select 10 exemplar images for each test images and calculate the final score using a multi-sample voting strategy.

We report the performance of various methods in our experiments, encompassing both the baseline method and different versions of our proposed methods. They are as follows :

$\bullet$  \bm{$F+R$} (Baseline) : Corresponds to the NR-AIGCIQA method proposed in \cite{yuan2023pkui2iqa}. This method first utilizes a backbone network to extract features and then uses a regression network composed of two fully connected layers to regress the predicted scores.

$\bullet$  \bm{$F+R+PS$} : Adding the patches sampling strategy on top of the baseline.

$\bullet$  \bm{$F+R+CR$} : Adding the contrastive regression framework on top of the baseline.

$\bullet$  \bm{$F+R+PSCR$} : Adding the patches sampling-based contrastive regression framework on top of the baseline.

\begin{table}[t]
\centering
\caption{Comparisons of performance with existing methods on three mainstream AIGCIQA database. * indicts our implementation.}
\label{tab:overall_table}

\begin{subtable}{.45\linewidth}
\centering
\caption{Comparisons of performance with existing methods on the AGIQA-1K database.}
\begin{tabular}{@{}l|cc@{}}
\toprule
\multirow{2}{*}{Method} & \multicolumn{2}{c}{MOS} \\ 
\cline{2-3}
& SRCC     & PLCC        \\ 
\hline
ResNet50\cite{zhang2023perceptual}      & 0.6365     & 0.7323        \\
StairIQA\cite{zhang2023perceptual}      & 0.5504     & 0.6088        \\
MGQA\cite{zhang2023perceptual}     & 0.6011     & 0.6760        \\
\hline
ResNet50*     & 0.8219     & 0.8255        \\
\hline
PS(ours)      & \textbf{0.8437}     & \textbf{0.8476}     \\
CR(ours)      & 0.8432     & 0.8403     \\
PSCR(ours)      & 0.8430     & 0.8403     \\
\bottomrule
\end{tabular}
\label{tab:sub_table1}
\end{subtable}
\begin{subtable}{.45\linewidth}
\centering
\caption{Comparisons of performance with existing methods on the AGIQA-3K database.}
\begin{tabular}{@{}l|cc@{}}
\toprule
\multirow{2}{*}{Method} & \multicolumn{2}{c}{MOS\_quality} \\ 
\cline{2-3}
& SRCC     & PLCC        \\ 
\hline
DBCNN\cite{AGIQA-3K}      & 0.8207     & 0.8759        \\
CLIPIQA\cite{AGIQA-3K}    & 0.8426     & 0.8053        \\
CNNIQA\cite{AGIQA-3K}     & 0.7478     & 0.8469        \\
HyperNet\cite{AGIQA-3K}   & 0.8355     & 0.8903        \\
\hline
PS(ours)      & 0.8314     & 0.8974     \\
CR(ours)      & 0.8477     & 0.9004     \\
PSCR(ours) & \textbf{0.8498}     & \textbf{0.9059}        \\
\bottomrule
\end{tabular}
\label{tab:sub_table2}
\end{subtable}

\begin{subtable}{.9\linewidth}
\centering
\caption{Comparisons of performance with existing methods on the AIGCIQA2023 database.}
\begin{tabular}{@{}l|cc|cc|cc@{}}
\toprule
 \multirow{2}{*}{Method} & \multicolumn{2}{c|}{Quality} & \multicolumn{2}{c|}{Authenticity} & \multicolumn{2}{c}{Correspondence} \\ \cline{2-7}  
 & SRCC & PLCC  & SRCC & PLCC & SRCC & PLCC  \\ 
\hline
CNNIQA\cite{wang2023aigciqa2023}   & 0.7160   & 0.7937    & 0.5958      & 0.5734     & 0.4758   & 0.4937    \\
VGG16\cite{wang2023aigciqa2023}    & 0.7961   & 0.7973    & 0.6660      & 0.6807     & 0.4580        & 0.6417   \\
VGG19\cite{wang2023aigciqa2023}    & 0.7733  & 0.8402    & 0.6674      & 0.6565   & 0.5799  & 0.5670  \\
ResNet18\cite{wang2023aigciqa2023}   & 0.7583 & 0.7763 & 0.6701 & 0.6528 & 0.5979 & 0.5564    \\
ResNet34\cite{wang2023aigciqa2023}   & 0.7229 & 0.7578 & 0.5998 & 0.6285 & 0.7058 & 0.7153    \\
\hline
VGG16*   & 0.8066   & 0.8225   & 0.7194   & 0.7106     & 0.6989     & 0.6883         \\
VGG19*   & 0.8071   & 0.8309  & 0.7127   & 0.7091     & 0.6862     & 0.6770     \\
ResNet18*   & 0.8005  & 0.8220  & 0.7259    & 0.7197  & 0.6997  & 0.6953   \\
\hline
PS(ours)   & 0.8319 & 0.8522 & 0.7501 & 0.7443 & 0.7298 & 0.7218 \\
CR(ours)   & \textbf{0.8389} & 0.8582 & 0.7585 & 0.7476 & 0.7464 & 0.7379    \\
PSCR(ours)   & 0.8371  & \textbf{0.8588} & \textbf{0.7828} & \textbf{0.7750} & \textbf{0.7465} & \textbf{0.7397}    \\
\bottomrule
\end{tabular}

\label{tab:sub_table3}
\end{subtable}%
\end{table}

\subsection{Results}

\subsubsection{Results on AGIQA-1K database.}
Table 1(a) shows the performance of existing methods and
our methods on AGIQA-1K database. We first conduct benchmark experiments using the NR-AIGCIQA method proposed in \cite{yuan2023pkui2iqa}, employing ResNet50\cite{he2016resnet} as the feature extraction network. In our experiments, the performance of ResNet50 significantly surpass that reported in the original paper. We also present the performance of various methods we proposed on the AGIQA-1K database. The results demonstrate that incorporating our proposed methods leads to notable performance improvements. In terms of SRCC, our proposed PS, CR, and PSCR methods achieve performance improvements of 2.18\%, 2.13\%, and 2.11\% respectively, compared to the Baseline. In terms of PLCC, our proposed PS, CR, and PSCR methods attain improvements of 2.21\%, 1.48\%, and 1.48\% respectively compared to the Baseline. Notably, the PS method achieve new state-of-the-art under both SRCC and PLCC.

\begin{table}[t]
\centering
\caption{Effects of different image preprocessing operations. on three mainstream AIGCIQA database, including AGIQA-1K, AGIQA-3K, and AIGCIQA2023. ``Baseline" corresponds to the image preprocessing operation as shown in Fig.1(a); ``Nonoverlapping" refers to the image preprocessing operation as shown in Fig.1(b); ``Overlapping" denotes the patches sampling strategy employed in our study as shown in Fig.1(c).}
\label{tab:overall_table}

\begin{subtable}{.45\linewidth}
\centering
\caption{Effects of different image preprocessing operations on the AGIQA-1K database.}
\begin{tabular}{@{}l|cc@{}}
\toprule
\multirow{2}{*}{Method} & \multicolumn{2}{c}{MOS} \\ 
\cline{2-3}
& SRCC     & PLCC        \\ 
\hline
Baseline  & 0.8168   & 0.8249   \\
Nonoverlapping    & 0.8312 & 0.8309      \\
Overlapping    & 0.8437   & 0.8358       \\ 
\bottomrule
\end{tabular}
\label{tab:sub_table1}
\end{subtable}
\begin{subtable}{.45\linewidth}
\centering
\caption{Effects of different image preprocessing operations on the AGIQA-3K database.}
\begin{tabular}{@{}l|cc@{}}
\toprule
\multirow{2}{*}{Method} & \multicolumn{2}{c}{MOS\_quality} \\ 
\cline{2-3}
& SRCC     & PLCC        \\ 
\hline
Baseline  &  0.8201   & 0.8795       \\
Nonoverlapping    & 0.8267   & 0.8874  \\
Overlapping    & 0.8314   & 0.8885        \\ 
\bottomrule
\end{tabular}
\label{tab:sub_table2}
\end{subtable}

\begin{subtable}{.9\linewidth}
\centering
\caption{Effects of different image preprocessing operations on the AIGCIQA2023 database.}
\begin{tabular}{@{}l|cc|cc|cc@{}}
\toprule
 \multirow{2}{*}{Method} & \multicolumn{2}{c|}{Quality} & \multicolumn{2}{c|}{Authenticity} & \multicolumn{2}{c}{Correspondence} \\ \cline{2-7}  
 & SRCC & PLCC  & SRCC & PLCC & SRCC & PLCC  \\ 
\hline
Baseline    & 0.8005   & 0.8220   & 0.7259   & 0.7197  & 0.6997     & 0.6953     \\
Nonoverlapping     & 0.8008  & 0.8281  & 0.6962  & 0.6894   & 0.6776 & 0.6656   \\
Overlapping       & 0.8176  & 0.8387   & 0.7256   & 0.7141   & 0.7024     & 0.6883    \\ 
\bottomrule
\end{tabular}

\label{tab:sub_table3}
\end{subtable}%
\end{table}

\subsubsection{Results on AGIQA-3K database.}
Table 1(b) shows the performance of existing methods and
our methods on AGIQA-3K database. We report the performance of various methods we proposed on the AGIQA-3K database. Under SRCC, our proposed CR and PSCR methods achieve performance improvements of 0.51\% and 0.72\% respectively compared to the current best results, while our PS method do not surpass the current best result. Under PLCC, our proposed PS, CR, and PSCR methods all exceed the current best result, achieving performance improvements of 0.71\%, 1.01\%, and 1.56\% respectively. Notably, the PSCR method achieve new state-of-the-art under both SRCC and PLCC.

\subsubsection{Results on AIGCIQA2023 database.}
Table 1(c) shows the performance of existing methods and
our methods on AIGCIQA2023 database. We initially replicate the benchmark experiments using the NR-AIGCIQA method proposed in \cite{yuan2023pkui2iqa}, employing models such as VGG16/VGG19/ResNet50 as feature extraction networks. In our experiments, the performance of models like VGG16/VGG19/\\ResNet50 slightly exceeds that reported in the original paper. We also report the performance of various methods we proposed on the AIGCIQA2023 database. The results indicate significant performance improvements through the integration of our proposed methods. In terms of quality, compared to the current best results, our proposed PS, CR, and PSCR methods achieve performance improvements of 2.48\%, 3.18\%, and 3.00\% under SRCC, and 2.13\%, 2.73\%, and 2.79\% under PLCC, respectively. in terms of authenticity, our methods achieve improvements of 2.42\%, 3.26\%, and 5.69\% under SRCC, and 2.46\%, 2.79\%, and 5.53\% under PLCC, respectively; in terms of correspondence, compared to the current best results, our proposed PS, CR, and PSCR methods achieve performance improvements of 3.01\%, 4.67\%, and 4.68\% under SRCC, and 2.65\%, 4.26\%, and 4.44\% under PLCC, respectively. Notably, the PSCR method achieves new state-of-the-art in all aspects except for quality under SRCC, while the CR method achieves new state-of-the-art in terms of quality under SRCC.

\begin{table}[t]
\scriptsize
\centering
\caption{Effects of start-index on three mainstream AIGCIQA database, including AGIQA-1K, AGIQA-3K, and AIGCIQA2023.}
\label{tab:overall_table}

\begin{subtable}{1.\linewidth}
\centering
\caption{Effects of start-index for ResNet18.}
\begin{tabular}{l|c|c|c|c|c|c|c|c|c|c}
\toprule
\multirow{3}{*}{Start-index} & \multicolumn{2}{c|}{AGIQA-1K} & \multicolumn{2}{c|}{AGIQA-3K} & \multicolumn{6}{c}{AIGCIQA2023} \\
\cline{2-11}
& \multicolumn{2}{c|}{MOS} & \multicolumn{2}{c|}{MOS\_quality} & \multicolumn{2}{c|}{Quality} &\multicolumn{2}{c|}{Authenticity} & \multicolumn{2}{c}{Correspondence} \\
\cline{2-11}
& SRCC & PLCC &  SRCC & PLCC & SRCC & PLCC & SRCC & PLCC & SRCC & PLCC \\
\hline
$[0,150,288]$  & 0.8437   & 0.8358  & 0.8314   & 0.8885   & 0.8176   & 0.8387   & 0.7256   & 0.7141  & 0.7024     & 0.6883     \\
$[0,100,200,288]$   & 0.8395 & 0.8345   & 0.8374   & 0.8939   & 0.8187  & 0.8447  & 0.7123  & 0.7113   & 0.6966 & 0.6876   \\
\bottomrule
\end{tabular}
\label{tab:sub_table1}
\end{subtable}

\begin{subtable}{1.\linewidth}
\centering
\caption{Effects of start-index for InceptionV4.}
\begin{tabular}{l|c|c|c|c|c|c|c|c|c|c}
\toprule
\multirow{3}{*}{Start-index} & \multicolumn{2}{c|}{AGIQA-1K} & \multicolumn{2}{c|}{AGIQA-3K} & \multicolumn{6}{c}{AIGCIQA2023} \\
\cline{2-11}
& \multicolumn{2}{c|}{MOS} & \multicolumn{2}{c|}{MOS\_quality} & \multicolumn{2}{c|}{Quality} &\multicolumn{2}{c|}{Authenticity} & \multicolumn{2}{c}{Correspondence} \\
\cline{2-11}
& SRCC & PLCC &  SRCC & PLCC & SRCC & PLCC & SRCC & PLCC & SRCC & PLCC \\
\hline
$[0,213]$  & 0.8249   & 0.8297  & 0.8299   & 0.8987   & 0.8300   & 0.8520   & 0.7554   & 0.7495  & 0.7207     & 0.7062     \\
$[0,100,213]$   & 0.8364 & 0.8356   & 0.8287   & 0.8974   & 0.8319  & 0.8522  & 0.7501  & 0.7443   & 0.7298 & 0.7218   \\
\bottomrule
\end{tabular}
\label{tab:sub_table2}
\end{subtable}

\end{table}

\subsubsection{Effects of different image preprocessing operations.}
To further demonstrate the effectiveness of our proposed patches sampling strategy, we compare the performance of several common image preprocessing operations on three mainstream databases. We utilize ResNet18\cite{he2016resnet} as the feature extraction network for our experiments. The results are presented in Table 2. The experimental results indicate that methods based on patches sampling show clear advantages over the Baseline and our proposed patches sampling strategy significantly outperforms the other two methods as shown in Fig.1. We believe this is due to the fact that the patch-based methods can extract richer detail features, resulting in better performance. When the image patches are larger, the image preprocessing operation shown in Fig.1(b) results in greater information loss, rendering it less effective than our proposed method; conversely, when the image patches are smaller, our proposed method may introduce issues of information redundancy, indicating potential inapplicability in such scenarios.

\begin{table}[t]
\scriptsize
\caption{Ablation study for our proposed methods on three mainstream AIGCIQA database, including AGIQA-1K, AGIQA-3K, and AIGCIQA2023. The best performance results are marked in \textcolor{red}{RED} and the second-best performance results are marked in \textcolor{blue}{BLUE}.}
\centering
\begin{tabular}{l|c|c|c|c|c|c|c|c|c|c}
\toprule
\multirow{3}{*}{Method} & \multicolumn{2}{c|}{AGIQA-1K} & \multicolumn{2}{c|}{AGIQA-3K} & \multicolumn{6}{c}{AIGCIQA2023} \\
\cline{2-11}
& \multicolumn{2}{c|}{MOS} & \multicolumn{2}{c|}{MOS\_quality} & \multicolumn{2}{c|}{Quality} &\multicolumn{2}{c|}{Authenticity} & \multicolumn{2}{c}{Correspondence} \\
\cline{2-11}
& SRCC & PLCC &  SRCC & PLCC & SRCC & PLCC & SRCC & PLCC & SRCC & PLCC \\
\hline
VGG16 & 0.8128 & 0.8212 & 0.8162 & 0.8877 & 0.8066 & 0.8225 & 0.7194 & 0.7106 & 0.6989 & 0.6883 \\
VGG19 & 0.8154 & 0.8145 & 0.8177 & 0.8840 & 0.8071 & 0.8309 & 0.7127 & 0.7091 & 0.6862 & 0.6770 \\
ResNet18 & 0.8168 & 0.8249 & 0.8201 & 0.8795 & 0.8005 & 0.8220 & 0.7259 & 0.7197 & 0.6997 & 0.6953 \\
ResNet50 & 0.8219 & 0.8255 & 0.8205 & 0.8826 & 0.8179 & 0.8351 & 0.7307 & 0.7269 & 0.7172 & 0.7050 \\
InceptionV4 & 0.8093 & 0.8115 & 0.8110 & 0.8785 & 0.8265 & 0.8422 & 0.7430 & 0.7245 & 0.7109 & 0.7030 \\
\hline
VGG16(PS)       & 0.8296 & 0.8261 & 0.8177 & 0.8875 & 0.8234 & 0.8455 & 0.7349 & 0.7235 & 0.7013 & 0.6899 \\
VGG19(PS)       & 0.8274 & 0.8303 & 0.8110 & 0.8896 & 0.8161 & 0.8452 & 0.7331 & 0.7262 & 0.6976 & 0.6854 \\
ResNet18(PS)    & \textcolor{red}{0.8437} & 0.8358 & 0.8314 & 0.8885 & 0.8176 & 0.8387 & 0.7256 & 0.7141 & 0.7024 & 0.6883 \\
ResNet50(PS)   & \textcolor{blue}{0.8433} & \textcolor{red}{0.8476} & 0.8220 & 0.8847 & 0.8217 & 0.8474 & 0.7267 & 0.7245 & 0.7076 & 0.7017 \\
InceptionV4(PS) & 0.8364 & 0.8356 & 0.8287 & 0.8974 & 0.8319 & 0.8522 & 0.7501 & 0.7443 & 0.7298 & 0.7218 \\
\hline
VGG16(CR)       & 0.8261 & 0.8236 & 0.8326 & 0.8881 & 0.8243 & 0.8449 & 0.7306 & 0.7247 & 0.6963 & 0.6936\\
VGG19(CR)       & 0.8137 & 0.8171 & 0.8262 & 0.8883 & 0.8152 & 0.8438 & 0.7412 & 0.7378 & 0.6802 & 0.6746\\
ResNet18(CR)   & 0.8232 & 0.8254 & 0.8256 & 0.8874 & 0.8185 & 0.8353 & 0.7256 & 0.7204 & 0.7081 & 0.7021 \\
ResNet50(CR)    & 0.8432 & 0.8403 & 0.8341 & 0.8918 & 0.8272 & 0.8459 & 0.7311 & 0.7243 & 0.7067 & 0.6974 \\
InceptionV4(CR) & 0.8165 & 0.8221 & \textcolor{blue}{0.8477} & \textcolor{blue}{0.9004} & \textcolor{red}{0.8389} & \textcolor{blue}{0.8582} & \textcolor{blue}{0.7585} & \textcolor{blue}{0.7476} & \textcolor{blue}{0.7464} & \textcolor{blue}{0.7379} \\
\hline
VGG16(PSCR)      & 0.8378 & 0.8325 & 0.8184 & 0.8901 & 0.8239 & 0.8478 & 0.7354 & 0.7306 & 0.7114 & 0.7066 \\
VGG19(PSCR)      & 0.8084 & 0.8146 & 0.8088 & 0.8849 & 0.8162 & 0.8437 & 0.7091 & 0.7033 & 0.7036 & 0.6984 \\
ResNet18(PSCR)   & 0.8365 & 0.8336 & 0.8324 & 0.8948 & 0.8162 & 0.8412 & 0.7425 & 0.7338 & 0.7192 & 0.7102 \\
ResNet50(PSCR)   & 0.8430 & \textcolor{blue}{0.8403} & 0.8285 & 0.8934 & 0.8274 & 0.8505 & 0.7499 & 0.7460 & 0.7241 & 0.7197 \\
InceptionV4(PSCR) & 0.8281 & 0.8324 & \textcolor{red}{0.8498} & \textcolor{red}{0.9059} & \textcolor{blue}{0.8371} & \textcolor{red}{0.8588} & \textcolor{red}{0.7828} & \textcolor{red}{0.7750} & \textcolor{red}{0.7465} & \textcolor{red}{0.7397} \\
\bottomrule
\end{tabular}
\end{table}

\subsubsection{Effects of start-index.}
We test the effects of different start-indices on model performance, the experimental results are shown in Table 3. As we can see, different start-indices yield different gains and they each have their own advantages and disadvantages. When the stride is small, more detail features are extracted from the image patches, but this also increases information redundancy and training costs; conversely, when the stride is larger, fewer detail features are extracted, but this helps reduce information redundancy and training costs.

\subsubsection{Ablation Study.}
To demonstrate the effectiveness of our proposed methods, we conduct extensive ablation study on three mainstream AIGCIQA databases, including AGIQA-1K\cite{zhang2023perceptual}, AGIQA-3K\cite{AGIQA-3K}, and AIGCIQA2023\cite{wang2023aigciqa2023}. Five different backbone networks are selected for the experiments, including VGG16\cite{simonyan2014vgg}, VGG19\cite{simonyan2014vgg}, ResNet18\cite{he2016resnet}, ResNet50\cite{he2016resnet}, and InceptionV4\cite{szegedy2017inception}. The results are presented in Table 3. As an example, we compare the performance of our proposed methods with the Baseline under SRCC and PLCC while employing InceptionV4 for feature extraction, as shown in Fig.3. These results indicate that by integrating our proposed PS, CR, and PSCR methods into the Baseline model, performance improvements of varying degrees are achieved in most cases, thereby demonstrating the effectiveness of our proposed methods.

\section{Conclusion}
In this paper, We propose a patches sampling-based contrastive regression framework. As a versatile framework, it can be conveniently integrated into existing IQA methods. The contrastive regression framework not only leverages differences among various inputs to learn a better representation space but also overcomes the limitation of previous NR-IQA methods by using exemplar AIGIs as references. Furthermore, we propose a patches sampling strategy to avoid geometric distortion and information loss. Extensive experiments on three mainstream AIGCIQA databases have demonstrated the effectiveness of our proposed PSCR framework. However, despite the effectiveness of our proposed method, the patches sampling strategy we propose which employs a sliding window to obtain overlapping image patches introduces new issues such as information redundancy, disruption of the overall image structure, and loss of global information, \textit{etc}. In future research, we will focus on how to balance the introduction of this patches sampling strategy and the problems it entails.

\clearpage
\bibliographystyle{plain}   
\bibliography{PSCR} 
\end{document}